%% file: main_split.tex
\documentclass[letterpaper,10pt,conference]{materials/ieeeconf}

\IEEEoverridecommandlockouts
\usepackage{fontspec}
\newfontfamily\ieeetimes[NFSSFamily=ptm]{texgyretermes-regular.otf}
\usepackage{amsmath,amssymb,bm}
\usepackage{booktabs,multirow,tabularx}
\usepackage{graphicx}
\usepackage{cite}
\usepackage{xcolor}
\usepackage{xspace}
\usepackage{microtype}
\usepackage{hyperref}

\hypersetup{
  hidelinks,
  pdftitle={MINT: A Unified Model for World-Space Camera and Hand Motion Estimation from Scalable Egocentric Pipeline Supervision},
  pdfauthor={Zijie Zhu, Weiren Cai, Yizhou Wang, Zhenjie Yang, Yide Liu, Jiahao Chen, Guanqi He}
}

\newcommand{\method}{\textsc{MINT}\xspace}
\newcommand{\pipeline}{\textsc{EgoPipeline}\xspace}
\newcommand{\framework}{\textsc{MINT}\xspace}
\newcommand{\todoresult}[1]{}
\newcommand{\todocite}[1]{}
\newcommand{\placeholderfigure}[2]{}

\newcommand{\missingfigurebox}[1]{%
  \fbox{\parbox[c][0.14\textheight][c]%
    {\dimexpr\linewidth-2\fboxsep-2\fboxrule\relax}{%
    \centering\ttfamily\scriptsize
    MISSING FIGURE\\[4pt]\detokenize{#1}}}}
\let\origincludegraphics\includegraphics
\renewcommand{\includegraphics}[2][]{%
  \IfFileExists{#2}{\origincludegraphics[#1]{#2}}{\missingfigurebox{#2}}}

\makeatletter
\renewcommand{\thetable}{\arabic{table}}
\renewcommand{\fnum@table}{Table~\thetable}
\long\def\@makecaption#1#2{%
  \ifx\@captype\@IEEEtablestring%
    \@IEEEtablecaptionsepspace%
  \else%
    \@IEEEfigurecaptionsepspace%
  \fi%
  \setbox\@tempboxa\hbox{\footnotesize #1.~~ #2}%
  \ifdim \wd\@tempboxa >\hsize%
    \setbox\@tempboxa\hbox{\footnotesize #1.~~ }%
    \parbox[t]{\hsize}{\footnotesize\noindent\unhbox\@tempboxa#2}%
  \else%
    \hbox to\hsize{\footnotesize\hfil\box\@tempboxa\hfil}%
  \fi%
  \ifx\@captype\@IEEEtablestring\vskip 3pt\fi}
\makeatother

\makeatletter
\newcommand{\spanningcaption}[1]{%
  \def\@captype{figure}%
  \refstepcounter{figure}%
  \@makecaption{\fnum@figure}{#1}%
}
\makeatother

\title{\LARGE\bfseries MINT: A Unified Model for World-Space Camera and Hand Motion Estimation from Scalable Egocentric Pipeline Supervision}
\author{%
  Zijie Zhu$^{1,3,4}$, Weiren Cai$^{3}$, Yizhou Wang$^{1,3}$, Zhenjie Yang$^{4}$,~%
  Yide Liu$^{3,5}$, Jiahao Chen$^{3,*}$, and Guanqi He$^{2,3,*}$%
  \thanks{$^{1}$ShanghaiTech University, $^{2}$Tsinghua University, $^{3}$Wuji Technology, $^{4}$The University of Hong Kong, and $^{5}$Zhejiang University.}%
  \thanks{$^{*}$Corresponding authors.}%
}

\begin{document}

\IEEEaftertitletext{%
  \vspace{0.5\baselineskip}%
  \centerline{\includegraphics[width=\textwidth]{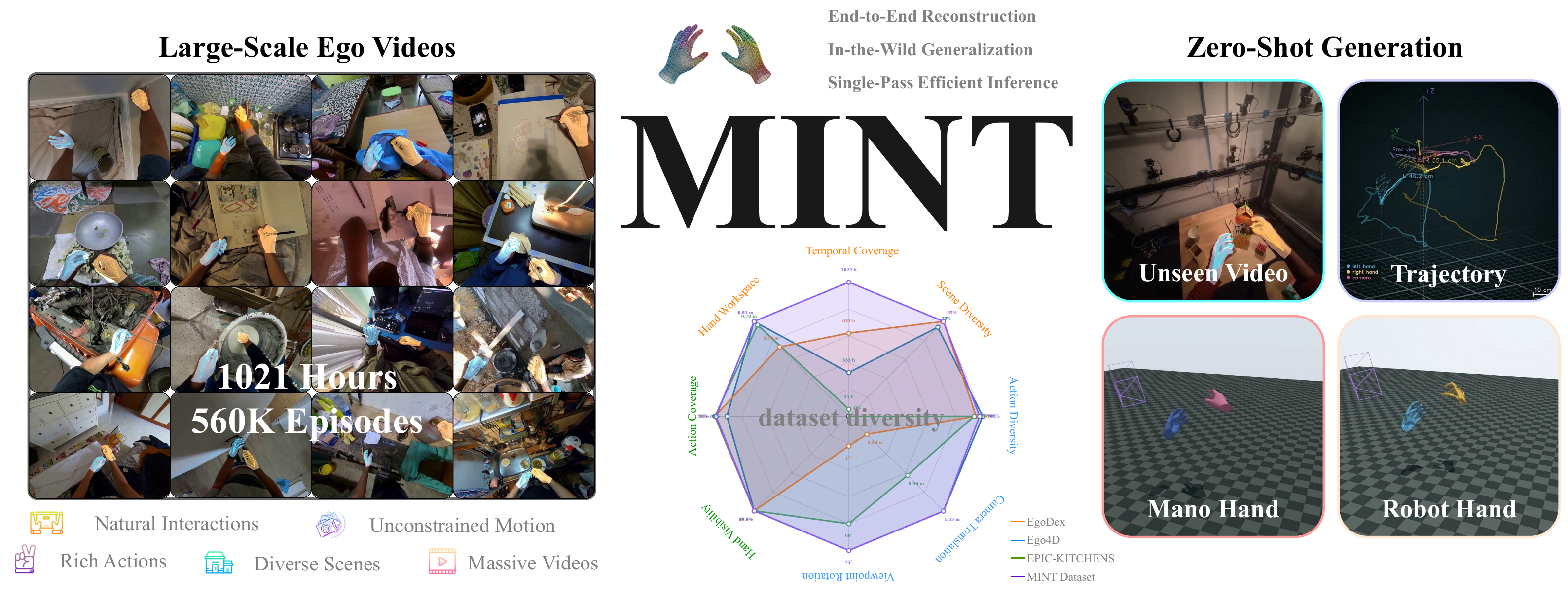}}%
  \spanningcaption{\framework{} converts ordinary egocentric video into world-space motion supervision, and \method{} recovers that motion with a single feed-forward model. Top: \pipeline{} labels $1{,}021$ hours of public egocentric video drawn from Ego4D~\cite{grauman2022ego4d}, EPIC-KITCHENS~\cite{damen2018epickitchens} and EgoDex~\cite{hoque2025egodex} (center) with camera trajectories and bimanual MANO~\cite{romero2017mano} states; each pair shows an input frame and the recovered hands reprojected onto it. Bottom: trained on that supervision, without any data from the evaluation domains, \method{} transfers zero-shot to unseen domains.}%
  \label{fig:teaser}%
  \vspace{0.5\baselineskip}%
}

\maketitle
\thispagestyle{empty}
\pagestyle{empty}

\input{sections/0_abs}

\input{sections/1_intro}

\input{sections/2_related}

\input{sections/3_pipeline}

\input{sections/4_method}

\input{sections/5_experiment}

\input{sections/7_conclusion}

\input{sections/8_bib}
\end{document}

%% file: sections/0_abs.tex
\begin{abstract}
% Activity understanding, robot imitation, and augmented reality require camera and bimanual motion in world coordinates. Existing systems recover this state by cascading camera calibration, monocular depth estimation, SLAM, hand reconstruction, and trajectory cleanup. Such pipelines repeatedly encode the same video, and they combine camera and hand trajectories only during post-processing, making deployment and maintenance expensive. We introduce \framework{} (Minting IN-the-Wild Trajectories) and its foundation model, \method. To our knowledge, \method is the first unified model to recover complete world-space bimanual trajectories end to end from egocentric RGB video. A shared spatiotemporal representation feeds a LingBot-Map camera-extrinsics head, an independent field-of-view head, a camera-frame MANO head, and a per-frame hand-presence head; explicit rigid transformations then produce world-space hand motion. We train \method in two stages: pretraining on structured pseudo-labels generated by the fully open-source \pipeline{} system, followed by fine-tuning on a small set of high-precision camera--hand annotations. We release the model weights, training and inference code, the complete labeling pipeline, and a rigorously filtered 1,021-hour structured egocentric dataset. The resulting model amortizes a distributed, multi-stage reconstruction pipeline into a single shared forward pass. We evaluate zero-shot generalization, world-space reconstruction, and end-to-end data-production efficiency on public benchmarks.

Recovering camera and hand motion in world coordinates from egocentric video is a key capability for activity understanding, robot learning, and augmented reality. Existing systems typically decompose this problem into separate stages for camera motion, depth estimation, hand reconstruction, and trajectory refinement, resulting in substantial computational overhead and preventing the joint modeling of camera and hand motion. We introduce \method(Minting IN-the-Wild Trajectories), a foundation model for world-space hand motion reconstruction from ego-centric RGB video. From a single shared spatiotemporal video representation, \method jointly predicts the camera trajectory, field of view (FoV), camera-frame hand states, and per-frame hand observability, and then produces world-space hand motion via explicit coordinate transformations.
Training such a model at scale is challenging, since paired world-space camera and hand annotations are scarce. We therefore develop an open-source labeling  \pipeline{} that converts large collections of public egocentric videos into structured camera-and-hand trajectory supervision. MINT is first pretrained on these large-scale pseudo-labels and then fine-tuned on a small set of high-quality camera-and-hand annotations. 
Across public benchmarks \method{} approaches state-of-the-art accuracy without seeing either benchmark in training, reaching $0.945$ frame accuracy, $13.646$\,mm PA-MPJPE-p and $55.058$\,px EPE-p for camera-frame bimanual reconstruction on HOT3D, $4.690$\,mm RPE-T and $0.284^{\circ}$ RPE-R for camera trajectory, and a $3.67\times$ end-to-end speedup over the labeling pipeline that supervises it. We release the model, training and inference code, labeling pipeline, and a curated 1,021-hour egocentric trajectory dataset.
\end{abstract}


%% file: sections/1_intro.tex
\section{Introduction}
\label{sec:introduction}

Egocentric video records rich human motion and hand--object interactions, and recovering their world-space motion provides an important source of motion supervision for embodied AI training~\cite{li2025vitra}, including robot imitation learning~\cite{kareer2025egomimic}. Yet accurate world-space supervision remains scarce. Capturing such supervision typically requires specialized sensing, precise calibration, synchronized devices, or costly motion reconstruction, making it difficult to obtain large-scale, high-quality world-space supervision from publicly available egocentric video.

Existing reconstruction methods decompose egocentric motion into separate modules for hand detection, motion reconstruction, and camera trajectory estimation. Such pipelines can recover individual sequences, but typically require complex optimization procedures and substantial engineering effort. Hand estimation depends on a detector and can fail under severe occlusion, while errors in cascaded reconstruction propagate across stages~\cite{ye2023slahmr,wang2024tram}, and representations are difficult to share across tasks. Meanwhile, large-scale 3D reconstruction models provide a basis for jointly modeling camera and hand motion. Their pretraining captures long-range temporal correspondence, camera motion, and scene geometry within a shared spatio-temporal representation. We therefore ask whether such a unified geometric representation can support both camera and hand motion reconstruction.

We present \method{}, a unified model for recovering world-space camera and bimanual hand motion from egocentric RGB video. \method{} adapts a pretrained geometric reconstruction model to egocentric data and predicts camera extrinsics, horizontal and vertical fields of view, left- and right-hand MANO~\cite{romero2017mano} states, and per-frame hand observability from a shared representation. It requires no external hand detector, motion infiller, or test-time optimization, and no dense depth or point-map intermediate. For each $32$-frame window, all outputs are obtained in a single encoder forward pass, while videos of arbitrary length are covered by chaining overlapping windows (Sec.~\ref{sec:backbone}), enabling unified motion reconstruction.

Real-world egocentric videos contain diverse scenes and hand motion, but lack accurate world-space hand supervision. We therefore build \pipeline{} to produce motion supervision at scale from ordinary video. We process 1,729 hours from Ego4D~\cite{grauman2022ego4d}, EPIC-KITCHENS~\cite{damen2018epickitchens}, and EgoDex~\cite{hoque2025egodex} into 1,021 hours of world-space camera and hand motion supervision (Fig.~\ref{fig:teaser}). Because the camera trajectories in this supervision inherit the scale ambiguity of monocular reconstruction~\cite{yin2023metric3d}, we introduce a second training stage that adapts the camera head alone using a human egocentric dataset captured with high-precision devices. The pseudo-labels provide broad coverage, while this dataset provides precise absolute scale.

Neither HOT3D~\cite{banerjee2025hot3d} nor ARCTIC~\cite{fan2023arctic} is included in the training data; both are therefore evaluated in the zero-shot setting. On HOT3D, \method{} approaches state-of-the-art accuracy on all nine metrics we report, remaining close to a model trained on that benchmark. For camera trajectory estimation, \method{} achieves the lowest relative translation error on ARCTIC.

Our contributions are:

\begin{itemize}
  \item \textbf{Unified egocentric reconstruction.} \method{} jointly recovers camera state, field of view, bimanual hand state, and hand observability from a unified spatio-temporal geometric representation. Each $32$-frame window requires a single encoder forward pass, with no detector, motion infiller, or test-time optimization.
  \item \textbf{Scalable egocentric data production.} \pipeline{} converts large-scale ordinary egocentric video into structured world-space motion supervision and combines broad, automatically reconstructed data with a small amount of precise geometric supervision.
  \item \textbf{Large-scale structured egocentric data.} We provide a broad corpus of reconstructed egocentric motion containing camera trajectories, bimanual hand state, world-space motion, hand observability, and action descriptions.
\end{itemize}

%% file: sections/2_related.tex
\section{Related Work}
\label{sec:related-work}

\subsection{Egocentric Video Understanding}

Large-scale egocentric video datasets, including Ego4D~\cite{grauman2022ego4d}, EPIC-KITCHENS~\cite{damen2018epickitchens}, Ego-Exo4D~\cite{grauman2024egoexo4d}, and EgoDex~\cite{hoque2025egodex}, provide important resources for learning human actions and visual-language representations from diverse real-world activities. ViTRA~\cite{li2025vitra} further demonstrates the value of large-scale real-life human videos for pretraining, covering diverse tasks, objects, and environments and improving zero-shot action prediction. Extracting 3D hand-motion supervision from such videos typically requires multiple stages, including camera calibration, depth estimation, visual SLAM, hand reconstruction, and trajectory processing, resulting in a complex data production pipeline with substantial engineering effort for optimization and deployment.

\begin{figure*}[t]
  \centering
  \includegraphics[width=\textwidth]{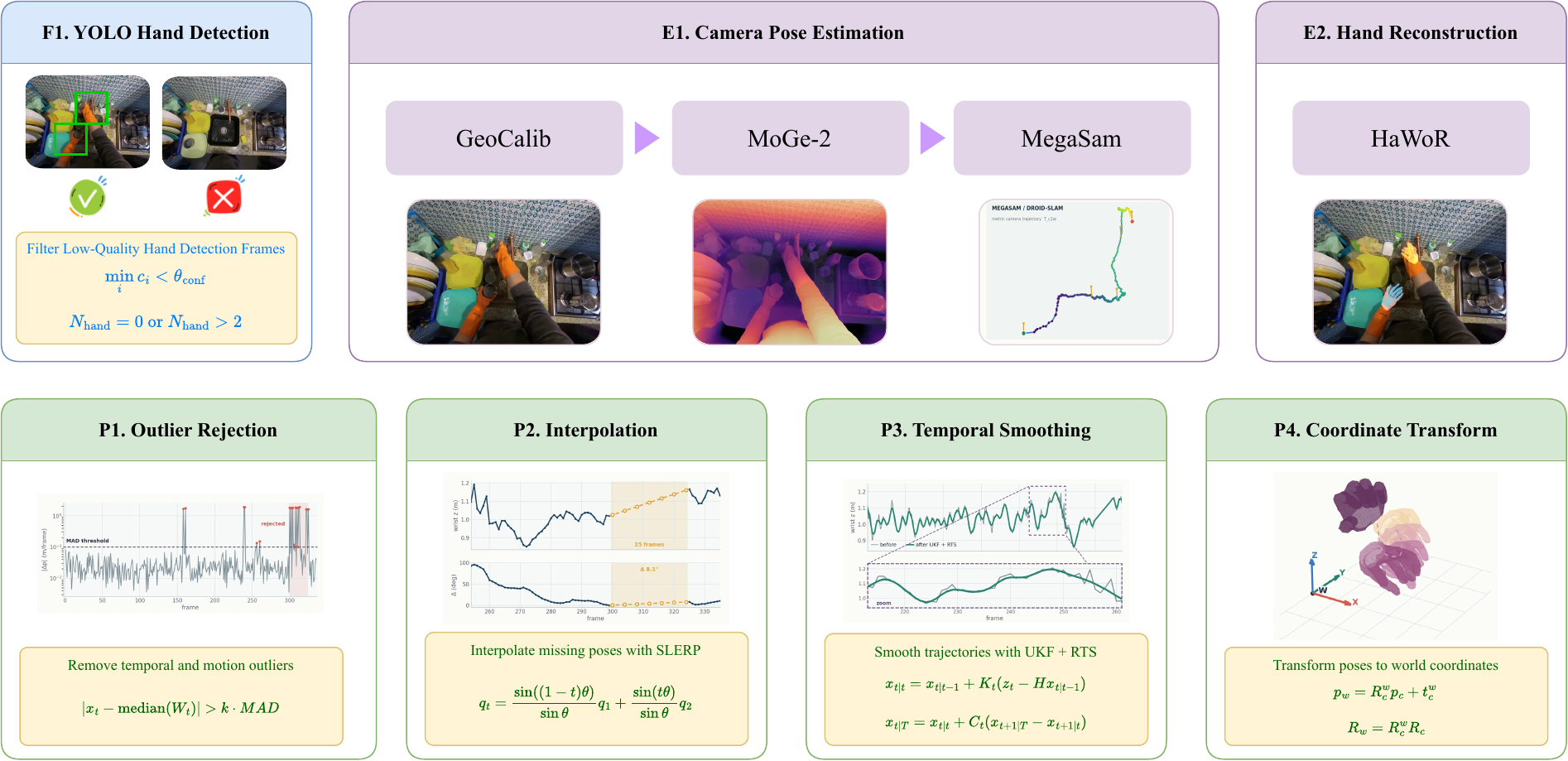}
  \caption{Overview of \pipeline. Egocentric videos are first pre-filtered by hand detection. The retained clips are then processed by GeoCalib for camera intrinsics, MoGe-2 for monocular depth, MegaSaM for camera pose, and HaWoR for bimanual hand reconstruction. A final post-processing stage stabilizes the recovered trajectories and composes the camera-frame hand states with the camera trajectory to produce world-space bimanual motion supervision.}
  \label{fig:ego-pipeline}
\end{figure*}

\subsection{Video-Based Hand Pose Estimation}

MANO~\cite{romero2017mano} provides a generic low-dimensional parameterization from images to hand pose and shape. Monocular hand reconstruction methods mainly recover hand pose and shape by estimating MANO~\cite{romero2017mano} parameters from RGB input. HaMeR~\cite{pavlakos2024hamer} employs a pretrained ViT with large-scale training data to improve generalization to real-world images, while WiLoR~\cite{potamias2025wilor} combines hand localization and 3D reconstruction for single-frame hand-state estimation~\cite{valassakis2024handdgp}. Video-based methods further introduce temporal modeling. Dyn-HaMR~\cite{yu2025dynhamr} incorporates a two-hand interaction motion prior~\cite{duran2024hmp} with test-time optimization, while HaWoR~\cite{zhang2025hawor} combines camera-frame hand estimation, egocentric SLAM trajectory estimation, and motion infilling based on monocular metric depth~\cite{yin2023metric3d}. ViDiHand~\cite{wang2026vidihand} exploits the spatiotemporal prior of a pretrained video diffusion model to jointly estimate hand presence and temporally stable 3D hand pose without an explicit detector or test-time optimization, while the reconstructed hand states remain in the camera frame.

The 3D supervision used by these methods is primarily obtained from dedicated sensing systems and staged laboratory captures, including the multi-depth-camera setup of DexYCB~\cite{chao2021dexycb}, the head-mounted system of HOT3D~\cite{banerjee2025hot3d}, and the multi-view marker-based motion capture system of ARCTIC~\cite{fan2023arctic}. These systems provide accurate 3D annotations but rely on specialized acquisition setups, with limited coverage of hand motions, object interactions, and real-world scenes. Egocentric body-and-hand estimation faces the same supervision bottleneck~\cite{yi2024egoallo}.

\subsection{Feed-Forward 3D Reconstruction}

Camera and scene reconstruction has increasingly moved from per-video optimization toward learned feed-forward estimation~\cite{wang2024dust3r,leroy2024mast3r}. DROID-SLAM~\cite{teed2021droidslam} combines deep networks with differentiable optimization to jointly estimate camera poses and dense geometry, while MegaSaM~\cite{li2025megasam} incorporates learned geometric representations into SLAM for dynamic and casually captured videos.

Recent feed-forward methods directly predict camera and scene geometry from images or videos~\cite{wang2025cut3r,zhang2025monst3r}. VGGT~\cite{wang2025vggt} uses cross-view attention to estimate camera parameters and dense geometry from multiple frames, while $\pi^3$~\cite{wang2025pi3} employs permutation-equivariant modeling to predict affine-invariant camera poses and scale-invariant local point maps. LingBot-Map~\cite{chen2026lingbotmap} extends feed-forward reconstruction to long streaming videos by aggregating historical observations into a spatiotemporal geometric representation, enabling continuous estimation of camera trajectories and scene geometry over long sequences. These works establish a feed-forward paradigm for geometric reconstruction from long video sequences. Our work extends this paradigm to egocentric hand reconstruction by jointly modeling camera and bimanual motion and recovering temporally consistent 3D hand motion in a unified world coordinate system.

\begin{figure*}[t]
  \centering
  \includegraphics[width=\textwidth]{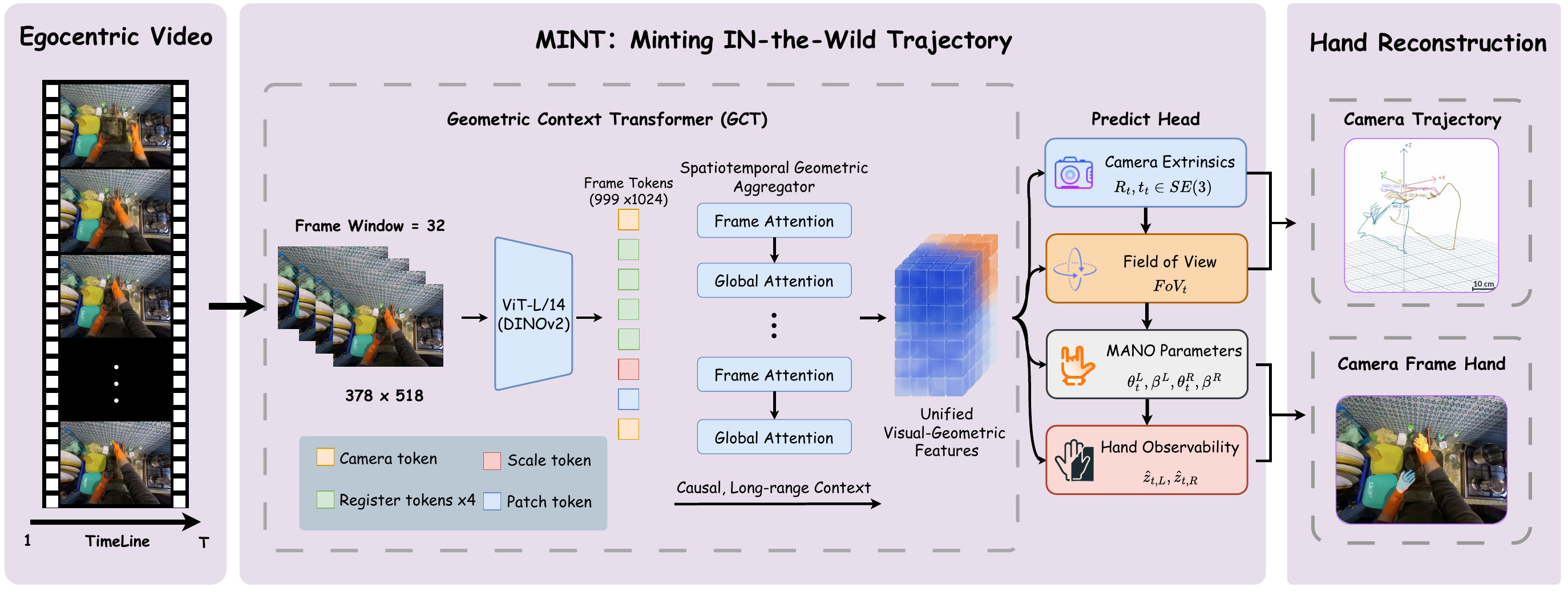}
  \caption{Architecture of \method. Each window of 32 egocentric frames is encoded at $378\times518$ by a DINOv2 ViT-L/14 patch embedding, and the resulting frame tokens are processed by the spatiotemporal geometric aggregator of GCT, which alternates frame attention and global attention so that every frame of the window is encoded with bidirectional temporal context. The unified visual-geometric features carry camera, register, scale, and patch tokens, and feed four prediction heads for camera extrinsics, field of view, bimanual MANO parameters, and hand observability. Composing the predicted camera trajectory with the camera-frame hand states yields world-space bimanual reconstruction.}
  \label{fig:model-architecture}
\end{figure*}

%% file: sections/3_pipeline.tex
\section{EgoPipeline Structured Supervision}
\label{sec:pipeline-supervision}

Existing egocentric hand reconstruction and hand--object interaction datasets are predominantly collected in controlled environments with predefined tasks, limiting the diversity of bimanual interactions and real-world conditions they capture~\cite{banerjee2025hot3d,chao2021dexycb,fan2023arctic}. In contrast, publicly available egocentric videos cover diverse activities, objects, environments, and interactions. To exploit this diversity for large-scale 3D motion learning, we collect 1,729 hours of egocentric video spanning open-ended daily activities, fine-grained bimanual manipulation on tabletops, and structured kitchen interactions. These videos contain visual and motion variations that are difficult to reproduce in controlled capture, including complex backgrounds, dynamic object interactions, severe hand occlusions, variations in hand scale, and illumination conditions.

To extract 3D motion supervision from these videos, we develop a first-person video processing pipeline inspired by the data construction paradigm of ViTRA~\cite{li2025vitra}, illustrated in Fig.~\ref{fig:ego-pipeline}. We first compress and encode the videos for improved I/O throughput and use high-confidence YOLO detections to filter clips with no detected hands for more than 1\,s or more than two detected hands. For the remaining clips, we estimate camera intrinsics with GeoCalib~\cite{veicht2024geocalib}, undistort the frames, and inject the monocular metric depth prior~\cite{yin2023metric3d} from MoGe-2~\cite{wang2025moge2} into MegaSaM~\cite{li2025megasam} or DROID-SLAM~\cite{teed2021droidslam} to recover metric-scale camera trajectories. In parallel, HaWoR~\cite{zhang2025hawor} reconstructs both hands and estimates per-frame motion states in the camera coordinate system, with each hand represented by a 6D wrist pose and MANO~\cite{romero2017mano} parameters. We then combine the camera-space hand states with the recovered camera trajectory to obtain world-space bimanual motion.

Because HaWoR relies on YOLO-based hand detections, reconstruction failures and intermittent detections can produce outliers, abrupt trajectory changes, and short missing intervals. We therefore apply unified post-processing with outlier removal, missing-interval interpolation, and UKF temporal filtering to stabilize the recovered trajectories and provide more reliable motion supervision. Finally, we construct 1,021 hours of effective data with world-coordinate 3D bimanual motion supervision.

Despite enabling large-scale supervision generation, the pipeline relies on multiple heterogeneous models for hand detection, camera calibration, depth estimation, SLAM, and hand reconstruction. This design incurs substantial computational and engineering overhead, while errors from different stages can propagate through the reconstruction process. We therefore propose \method{}, a unified feed-forward egocentric 3D reconstruction framework that jointly models camera and bimanual motion directly from video. \method{} learns their coupled geometry and predicts camera and hand motion in a common world coordinate system.

%% file: sections/4_method.tex
\section{Method}
\label{sec:method}

\subsection{Overview}
\label{sec:overview}

Monocular hand reconstruction aims to recover 3D hand pose and motion from egocentric videos. Existing methods typically rely on a pipeline of hand detection, camera tracking, hand fitting, and post-processing, which requires substantial engineering effort, accumulates errors across stages, and limits overall processing throughput. We therefore leverage a pretrained Geometric Context Transformer (GCT), the geometric encoder of LingBot-Map~\cite{chen2026lingbotmap} trained for long-range 3D reconstruction from video, to learn a unified visual-geometric representation for hand reconstruction. GCT encodes a window of frames in a single bidirectional spatiotemporal pass, and four task heads regress camera extrinsics, field of view (FoV), MANO parameters for both hands, and hand observability, as shown in Fig.~\ref{fig:model-architecture}. \method{} recovers camera and bimanual hand motion in the world coordinate frame from egocentric video with a single feed-forward model, without a hand detector, motion infiller, or test-time optimization. A window is processed in one pass, and videos longer than the window are covered by chaining windows as described in Sec.~\ref{sec:backbone}.

We adopt a two-stage training strategy. In the first stage, the visual-geometric representation and task heads are jointly trained on diverse egocentric datasets to establish coupled camera and hand reconstruction. In the second stage, while preserving the learned geometric representation and hand reconstruction capability, we adapt only the camera extrinsics head on a small in-house corpus whose camera trajectories are captured with high-precision devices. This curriculum transfers long-range 3D reconstruction priors to egocentric hand motion recovery while specializing camera estimation for accurate world-space annotation.

\subsection{Visual Geometry Grounded Backbone}
\label{sec:backbone}

Egocentric video interleaves articulated hand motion, hand--object contact, camera egomotion, and the surrounding 3D layout in a single stream, and none of them is fully recoverable from an isolated frame. To obtain temporally consistent geometric features, \method{} adopts the pretrained geometric context transformer (GCT) of LingBot-Map~\cite{chen2026lingbotmap}, a visual geometry grounded transformer~\cite{wang2025vggt}, as its shared spatiotemporal geometric encoder, in contrast to pairwise reconstruction backbones~\cite{wang2024dust3r,leroy2024mast3r}. Reconstruction training over a broad distribution of scenes has established in that model robust geometric priors for scene structure, viewpoint change, and trajectory consistency, and its aggregation of features along a video sequence also matches the temporal input of egocentric recordings more closely than models built for unordered image sets.

\paragraph{Geometric feature extraction}
We retain the original visual encoder and geometric aggregator so that these pretrained priors are inherited. Each frame is resized to $378\times518$ and encoded by a ViT-L/14 encoder, initialized from DINOv2~\cite{oquab2024dinov2} and pretrained jointly with the aggregator, into $999$ tokens of dimension $1024$. One camera token, four register tokens, and one scale token are prepended to every frame, so that frame-level global quantities are carried in dedicated slots. The sequence then alternates between frame attention and global attention,
\begin{equation}
  \bm{X}^{l}_{1:T}=
  G^{l}_{\mathrm{global}}
  \left(G^{l}_{\mathrm{frame}}(\bm{X}^{l-1}_{1:T})\right),
\end{equation}
where a frame module models the spatial structure inside a single image and a global module propagates geometric context along time, so that egomotion is inferred from cross-frame evidence rather than from single-frame visual cues. Features taken from several depths are aggregated into shared prediction features, which the camera and bimanual heads of Sec.~\ref{sec:heads} read from the same forward pass.

\paragraph{Long-sequence inference}
The global attention of GCT operates on a clip of fixed length and is trained at $T=32$ frames, so extending attention across a complete recording would let computation and memory grow with sequence length. We therefore segment a video of arbitrary length into overlapping sliding windows. Each window is encoded independently by the same shared network, the camera trajectories of adjacent windows are aligned and chained in $\mathrm{SE}(3)$, and the hand parameters and observability scores of the frames two windows share are fused. This design bounds the computation of a single pass to one fixed window while keeping the total cost approximately linear in the length of the video, which supports continuous geometric estimation over long sequences.

\subsection{Prediction Heads}
\label{sec:heads}

Through large-scale geometric reconstruction pretraining, GCT acquires strong priors for scene geometry, viewpoint change, and temporal motion, allowing its features to encode scene structure, viewpoint changes, and cross-frame motion. However, these generic geometric features do not directly specify the structured state required for hand reconstruction. Unlike rigid scene geometry, egocentric hand reconstruction must jointly recover wrist pose, articulated joint motion, and hand shape while handling occlusion and partial observability during hand--object interaction. We therefore introduce dedicated hand and camera prediction heads on top of these geometric features, mapping them to structured bimanual MANO~\cite{romero2017mano} states and camera states. The hand state is recovered in the camera frame and coupled with camera motion through a differentiable camera-to-world transformation, while a hand observability head suppresses unreliable supervision from partially observed hands.

For each hand $s\in\{L,R\}$, we parameterize its state as
\begin{equation}
  \hat{\bm{h}}_{t,s}=
  [\hat{\bm{\Lambda}}_{t,s},\hat{\bm{\Gamma}}_{t,s},
   \hat{\bm{\Theta}}_{t,s},\hat{\bm{\beta}}_{t,s}]
  \in\mathbb{R}^{109},
  \label{eq:hand-state}
\end{equation}
where $\bm{\Lambda}$, $\bm{\Gamma}$, $\bm{\Theta}$, and $\bm{\beta}$ denote wrist translation, wrist rotation, fifteen joint rotations, and hand shape, respectively. All rotations use the continuous six-dimensional representation, resulting in $218$ hand parameters for both hands at each frame.

The camera state is parameterized as
\begin{equation}
  \hat{\bm{c}}_{t}=[\hat{\bm{t}}_{t},\hat{\bm{q}}_{t},\hat{\bm{f}}_{t}]
  \in\mathbb{R}^{9},
  \label{eq:camera-state}
\end{equation}
comprising the camera translation $\bm{t}_t$, a unit quaternion $\bm{q}_t$ with rotation matrix $\bm{R}_t$ under the world-to-camera convention, and the vertical and horizontal fields of view $\bm{f}_t$.

\paragraph{Structured Hand Motion Prediction}
Egocentric hand motion follows a structured kinematic model, and frame-wise features alone are insufficient to reliably recover the full MANO state, particularly under rapid motion and frequent occlusion. We therefore directly decode the structured states of both hands from the spatio-temporal geometric features.

Each hand carries four learnable queries corresponding to wrist translation, wrist rotation, joint rotations, and hand shape. Let $\bm{X}_t$ denote the final-layer spatio-temporal tokens of frame $t$. Each query interacts with these tokens through
\begin{equation}
  \bm{Q}^{s}_{t,k}=
  \mathrm{CrossAttn}(\bm{q}^{s}_{k},\bm{X}_{t},\bm{X}_{t}),
  \quad k\in\{\Lambda,\Gamma,\Theta,\beta\}.
\end{equation}
The four resulting features are jointly mapped to the MANO state.

Frame-wise decoding alone does not guarantee temporal consistency, since rapid motion and occlusion can cause discontinuities across neighboring frames. We therefore condition the hand queries on the previous prediction and perform two refinement iterations. The first iteration predicts absolute rotations, while subsequent iterations predict a bounded axis-angle increment in the current local frame:
\begin{equation}
  \tilde{\bm{\omega}}=
  \frac{\bm{\omega}}{\lVert\bm{\omega}\rVert}\,
  \omega_{\max}\tanh\!\left(
  \frac{\lVert\bm{\omega}\rVert}{\omega_{\max}}\right),
  \qquad \omega_{\max}=30^{\circ}.
\end{equation}
The update is then applied as $\bm{R}^{(k+1)}=\bm{R}^{(k)}\exp([\tilde{\bm{\omega}}^{(k)}]_{\times})$, and mapped back to the continuous six-dimensional representation. The residual layers are zero-initialized, making the refinement module an identity mapping at the beginning of training.

\input{tables/tab_camera_hand}

\paragraph{Camera State and Hand--Camera Geometric Coupling}
The hand state is recovered in the camera frame, whereas the final hand motion must be expressed in a common world frame. The camera state therefore determines both camera motion and the mapping of the hand trajectory into world space.

We adopt the LingBot-Map~\cite{chen2026lingbotmap} camera head, which predicts camera pose through iterative refinement. The previous prediction is embedded and injected into a causal temporal trunk through adaptive layer normalization, which predicts the current update. We retain four refinement iterations and move the field of view out of the feedback path, reducing the iterative state from nine to seven dimensions.

The field of view is predicted by an independent branch:
\begin{equation}
  \hat{\bm{f}}_t=\mathrm{Softplus}
  \big(H_{\mathrm{FoV}}(\bm{X}^{\mathrm{cam}}_t)\big),
\end{equation}
where the camera token is first projected to $512$ dimensions and processed by two temporal blocks. The predicted extrinsics and fields of view are then concatenated into the camera state in Eq.~\eqref{eq:camera-state}.

Let $\bm{p}^{c}_{t,s}$ and $\bm{Q}^{c}_{t,s}$ denote the wrist translation and rotation matrix of hand $s$ in the camera frame. They are mapped to the world frame as
\begin{equation}
  \bm{p}^{w}_{t,s}=\bm{R}_t^{\top}(\bm{p}^{c}_{t,s}-\bm{t}_t),
  \quad
  \bm{Q}^{w}_{t,s}=\bm{R}_t^{\top}\bm{Q}^{c}_{t,s}.
  \label{eq:world-composition}
\end{equation}
Joint rotations and hand shape remain invariant under this transformation. Since the transformation is differentiable, the world-frame loss can jointly constrain the camera and hand predictions, explicitly coupling their geometry.

\paragraph{Hand Observability}
Hands in egocentric videos are frequently affected by object interaction, self-occlusion, and image-boundary truncation, making some frames unreliable for geometric supervision. We therefore introduce a lightweight hand observability head to determine whether each hand in the current frame can provide reliable supervision.

For each hand, a learnable query cross-attends to the current-frame patch tokens. The tokens are projected to $256$ dimensions, and the head predicts the left- and right-hand observability logits $\hat{z}_{t,L}$ and $\hat{z}_{t,R}$. The corresponding probabilities $\hat{a}_{t,s}$ determine whether that hand contributes to the hand reconstruction loss and world-frame loss.

\subsection{Training Objectives}
\label{sec:training}

The egocentric reconstruction pipeline provides high-quality hand motion from large-scale videos, including wrist trajectories, joint motion, and hand--object interactions, while covering a broad range of tasks, scenes, and action variations. These data provide large-scale supervision for learning hand motion and geometric relationships in egocentric videos.

However, the corresponding camera trajectories typically rely on monocular depth and SLAM. Although monocular SLAM captures relative frame-to-frame motion well, its scale ambiguity can introduce systematic errors in the absolute trajectory scale. We therefore use two training stages: the first learns stable hand--camera geometry from large-scale and diverse data, while the second uses camera motion captured with high-precision devices to correct the remaining scale and absolute pose errors.

\paragraph{Stage 1: Learning Hand--Camera Coupled Geometry}
The first stage does not directly fit an absolutely accurate camera trajectory. Instead, it exploits rich hand-motion supervision from large-scale egocentric data to learn a stable hand--camera geometric relationship. To improve generalization across tasks, actions, and scenes, we select clips with high scene and action diversity from Ego4D~\cite{grauman2022ego4d}, EgoDex~\cite{hoque2025egodex}, and EPIC-KITCHENS~\cite{damen2018epickitchens}, and initialize the geometric encoder with the pretrained GCT from LingBot-Map~\cite{chen2026lingbotmap}.

We jointly optimize four objectives: the camera motion loss $\mathcal{L}_{\mathrm{cam}}$, bimanual MANO loss $\mathcal{L}_{\mathrm{mano}}$, hand observability loss $\mathcal{L}_{\mathrm{obs}}$, and world-frame consistency loss $\mathcal{L}_{\mathrm{world}}$:
\begin{equation}
  \mathcal{L}_{\mathrm{stage1}}=
  \lambda_{\mathrm{cam}}\mathcal{L}_{\mathrm{cam}}+
  \lambda_{\mathrm{mano}}\mathcal{L}_{\mathrm{mano}}+
  \lambda_{\mathrm{obs}}\mathcal{L}_{\mathrm{obs}}+
  \lambda_{\mathrm{world}}\mathcal{L}_{\mathrm{world}}.
  \label{eq:total-loss}
\end{equation}
Through the differentiable transformation in Eq.~\eqref{eq:world-composition}, the world-frame loss jointly constrains the camera and hand predictions, allowing reliable hand-motion supervision to inform their geometric relationship.

The field of view and camera trajectory are predicted by separate heads, preserving the pretrained field-of-view capability while preventing camera-trajectory adaptation from interfering with it.

\paragraph{Stage 2: Camera Trajectory Correction}
The first stage establishes stable hand--camera geometry, but monocular SLAM can still leave errors in absolute scale and pose. We therefore perform a second stage using camera trajectories captured with high-precision devices.

We freeze the learned geometric encoder, hand, observability, and field-of-view modules, and optimize only the camera trajectory prediction. The objective consists of a translation term $\mathcal{L}_{t}$, a rotation term $\mathcal{L}_{R}$, and a temporal consistency term $\mathcal{L}_{\mathrm{temp}}$:
\begin{equation}
  \mathcal{L}_{\mathrm{stage2}}=
  \lambda_{t}\mathcal{L}_{t}+
  \lambda_{R}\mathcal{L}_{R}+
  \lambda_{\mathrm{temp}}\mathcal{L}_{\mathrm{temp}}.
  \label{eq:stage2-loss}
\end{equation}
Since hand reconstruction and hand--camera geometry are established in Stage 1, a small amount of high-precision camera trajectory data is sufficient to correct the absolute scale and pose while preserving the learned hand motion, hand--camera geometry, and field-of-view estimation.

%% file: tables/tab_camera_hand.tex
\begin{table*}[t]
  \centering
  \caption{\textbf{Main comparison on the ARCTIC and HOT3D datasets.} Best result per column in bold. $\uparrow$/$\downarrow$ indicate the direction of improvement. MPJPE-p and PA-MPJPE-p are in mm, EPE-p in px, GO-p in degrees, CT-p in m and Jitter in mm/frame$^2$. \method{} is evaluated in a zero-shot setting.}
  \label{tab:camera-hand}
  \scriptsize
  \setlength{\tabcolsep}{4pt}
  \renewcommand{\arraystretch}{0.95}
  \begin{tabular*}{\textwidth}{@{\extracolsep{\fill}}llccccccccc@{}}
    \toprule
      & & \multicolumn{3}{c}{Detection} & \multicolumn{2}{c}{3D Pose} & \multicolumn{3}{c}{Orient.\ \& Position} & Temporal \\
      \cmidrule(lr){3-5}\cmidrule(lr){6-7}\cmidrule(lr){8-10}\cmidrule(lr){11-11}
      & Method & FAcc $\uparrow$ & Recall $\uparrow$ & F1 $\uparrow$ & MPJPE-p $\downarrow$ & PA-MPJPE-p $\downarrow$ & EPE-p $\downarrow$ & GO-p $\downarrow$ & CT-p $\downarrow$ & Jitter $\downarrow$ \\
    \midrule
      \multirow{10}{*}{\rotatebox[origin=c]{90}{ARCTIC}}
      & InterWild~\cite{moon2023interwild} & 0.878 & 0.943 & 0.959 & 30.817 & 15.952 & 53.888 & 25.386 & 0.097 & 46.577 \\
      & HaMeR~\cite{pavlakos2024hamer} & 0.875 & 0.943 & 0.957 & 29.197 & 14.596 & 65.289 & 24.907 & 0.095 & 18.279 \\
      & Hamba~\cite{dong2024hamba} & 0.833 & 0.912 & 0.941 & 31.233 & 17.168 & 87.047 & 27.822 & 0.110 & 15.357 \\
      & WildHands~\cite{prakash2024wildhands} & 0.879 & 0.946 & 0.960 & 25.704 & 13.941 & \textbf{50.517} & 22.320 & \textbf{0.058} & 12.972 \\
      & OmniHands~\cite{lin2024omnihands} & 0.866 & 0.949 & 0.954 & 29.674 & 14.203 & 51.505 & 24.580 & 0.087 & 45.312 \\
      & WiLoR~\cite{potamias2025wilor} & \textbf{0.919} & 0.951 & 0.974 & \textbf{22.012} & \textbf{11.873} & 71.527 & \textbf{17.358} & 0.075 & 24.091 \\
      & Dyn-HaMR~\cite{yu2025dynhamr} & 0.842 & 0.918 & 0.951 & 27.904 & 17.017 & 85.723 & 25.951 & 0.121 & 12.840 \\
      & HaWoR~\cite{zhang2025hawor} & 0.700 & 0.817 & 0.895 & 45.357 & 26.375 & 158.062 & 43.325 & 0.149 & 19.789 \\
    \cmidrule(l){2-11}
      & \method{} & 0.918 & \textbf{0.957} & \textbf{0.978} & 26.527 & 15.961 & 55.038 & 19.332 & 0.069 & 12.959 \\
      & \method{} + UKF & 0.918 & \textbf{0.957} & \textbf{0.978} & 26.568 & 15.955 & 54.997 & 19.353 & 0.069 & \textbf{2.621} \\
    \midrule
      \multirow{10}{*}{\rotatebox[origin=c]{90}{HOT3D}}
      & InterWild~\cite{moon2023interwild} & 0.669 & 0.881 & 0.868 & 77.168 & 24.811 & 71.482 & 58.501 & 0.213 & 101.164 \\
      & HaMeR~\cite{pavlakos2024hamer} & 0.692 & 0.904 & 0.883 & 68.314 & 21.455 & 59.077 & 49.636 & 0.102 & 23.632 \\
      & Hamba~\cite{dong2024hamba} & 0.632 & 0.828 & 0.853 & 71.732 & 29.620 & 107.625 & 56.525 & 0.128 & 18.507 \\
      & WildHands~\cite{prakash2024wildhands} & 0.655 & 0.863 & 0.844 & 52.791 & 28.946 & 111.438 & 53.933 & 0.157 & 22.885 \\
      & OmniHands~\cite{lin2024omnihands} & 0.649 & 0.895 & 0.868 & 63.281 & 22.682 & 68.437 & 49.120 & 0.133 & 69.510 \\
      & WiLoR~\cite{potamias2025wilor} & 0.827 & 0.897 & 0.937 & 30.966 & 19.980 & 72.978 & 25.746 & \textbf{0.098} & 17.976 \\
      & Dyn-HaMR~\cite{yu2025dynhamr} & 0.614 & 0.811 & 0.802 & 74.214 & 38.201 & 171.617 & 43.851 & 0.571 & 44.942 \\
      & HaWoR~\cite{zhang2025hawor} & 0.348 & 0.499 & 0.654 & 71.396 & 66.031 & 327.294 & 79.350 & 0.262 & 23.872 \\
    \cmidrule(l){2-11}
      & \method{} & \textbf{0.945} & \textbf{0.983} & \textbf{0.953} & 29.926 & 13.656 & 55.117 & 21.099 & 0.196 & 12.057 \\
      & \method{} + UKF & \textbf{0.945} & \textbf{0.983} & \textbf{0.953} & \textbf{29.918} & \textbf{13.646} & \textbf{55.058} & \textbf{21.091} & 0.196 & \textbf{2.373} \\
    \bottomrule
  \end{tabular*}
\end{table*}

%% file: sections/5_experiment.tex
\section{Experiments}
\label{sec:experiments}

\subsection{Setup}
\label{sec:exp-setup}

\textbf{Datasets.} Stage~1 uses the 1,021 hours of \pipeline{} output from Sec.~\ref{sec:pipeline-supervision}, which contains no manually annotated hands and no ground-truth camera poses. Stage~2 trains the camera head alone on a small in-house corpus recorded with a single rig, using the camera trajectories that rig provides; no hand supervision is used in this stage. Neither stage sees HOT3D~\cite{banerjee2025hot3d} or ARCTIC~\cite{fan2023arctic}, on which all results below are therefore zero-shot.

\textbf{Metrics.} Standard hand pose metrics typically evaluate only successfully matched hands and therefore do not account for missed detections. We therefore use a coverage-aware evaluation in which missed hands are penalized rather than excluded from pose evaluation~\cite{wang2026vidihand}, and we group the metrics by what they measure. Detection is measured by frame accuracy (FAcc), recall and F1. Camera-frame hand reconstruction is measured by MPJPE-p and PA-MPJPE-p for articulated pose in 3D, by EPE-p for the reprojected 2D keypoint error, by GO-p for wrist orientation and by CT-p for hand placement, all computed under the coverage-aware protocol. For camera trajectories we report RPE-T and RPE-R for relative motion accuracy, and the arc-length ratio for scale deviation. Trajectories are evaluated over the full sequence and aligned with $SE(3)$ without scale fitting, so the reported errors retain the metric scale the model recovers on its own. RPE is computed per sequence as an RMSE and summarized by a mean and a median. Temporal quality is assessed by RPE for the camera and by Jitter for the hands. RPE is particularly relevant to \method{} because the model is trained with inter-frame motion increments, which makes it a direct measure of local motion consistency.

\textbf{Implementation.} \method{} is trained on $32$-frame clips at $378\times518$ resolution using AdamW with BF16 mixed precision. Stage~1 uses learning rates of $5\times10^{-5}$, $1\times10^{-4}$ and $5\times10^{-4}$ for the pretrained backbone, the camera and field-of-view heads and the hand and observability heads respectively, with an effective batch size of $1$. In stage~2 the aggregator, the field-of-view head and the hand and observability heads are frozen, and only the camera head is optimized at a learning rate of $5\times10^{-5}$, with an effective batch size of $4$. Both stages use a weight decay of $0.05$ and a $2{,}000$-step warmup.

\subsection{Camera-Frame Hands}
\label{sec:exp-camera-hand}

\textbf{Detection and pose.} Predicting both hands on every frame with an explicit observability score, rather than delegating to an external detector, is what makes \method{} reliable at finding hands under egocentric occlusion (Table~\ref{tab:camera-hand}). On HOT3D it lifts frame accuracy from $0.827$ for the strongest baseline to $0.945$ and recall from $0.904$ to $0.983$, so the share of frames carrying a missed hand falls from roughly one in six to one in twenty. On ARCTIC, recall and F1 are the highest of any method.

In pose accuracy \method{} scores well on both zero-shot benchmarks. On ARCTIC it stays behind WiLoR~\cite{potamias2025wilor} on the millimetre metrics while EPE-p comes within a few pixels of the best result, and on HOT3D it is the best method on every pose measure, reaching $29.918$\,mm MPJPE-p, $13.646$\,mm PA-MPJPE-p, $55.058$\,px EPE-p and $21.091^{\circ}$ GO-p. \method{} therefore maintains consistent performance across the two benchmarks, demonstrating strong cross-benchmark generalization.

The low EPE-p on HOT3D indicates accurate image-space localization, while CT-p reaches $0.069$\,m on ARCTIC and $0.196$\,m on HOT3D. These results indicate that the residual translation error primarily arises from metric depth estimation along the viewing ray, rather than from image-space localization.

\textbf{Temporal smoothness.} Egocentric hand trajectories are normally stabilized by a learned motion prior~\cite{duran2024hmp,ye2023slahmr} or by test-time optimization, both of which add inference cost, and \method{} needs neither to reach $12.959$ Jitter on ARCTIC, on par with the smoothest baseline, and $12.057$ on HOT3D, the best of any method. The bidirectional spatiotemporal encoder resolves each frame against its neighbors instead of smoothing an independently estimated sequence after the fact. The UKF of Sec.~\ref{sec:pipeline-supervision}, applied at inference, drives Jitter to $2.621$ and $2.373$, close to a fivefold reduction on both benchmarks. Smoothness is therefore already a property of the learned representation, and the filter sharpens it without disturbing the recovered geometry.

\subsection{Camera Trajectory}
\label{sec:exp-camera-traj}

\input{tables/tab_camera_traj}
\input{tables/tab_init}

On camera trajectory (Table~\ref{tab:camera-traj}) \method{} more than halves the relative translation error on ARCTIC, from $8.730$\,mm for the strongest baseline to $3.390$\,mm, and reaches $4.690$\,mm zero-shot on HOT3D, second only to MegaSaM. Absolute error is where the dedicated systems stay ahead, since a feed-forward pass over $32$-frame windows has neither loop closure nor global optimization and therefore accumulates drift over a full sequence.

The second stage acts on scale rather than on shape. Without it the model reaches an arc-length ratio of $0.466$ on HOT3D, so the recovered path is less than half the length of the true one, the signature of pseudo-labels whose scale comes from monocular depth. Stage~2 moves that ratio to $1.094$ and improves RPE-T from $8.750$ to $4.690$\,mm. On ARCTIC the same correction overshoots, taking the ratio from $0.755$ to $1.412$, so the recovered path now exceeds the true length even though RPE-T still improves from $3.530$ to $3.390$\,mm. Stage~2 therefore removes the systematic under-scaling that dominates on HOT3D, at the price of some over-scaling on a benchmark whose motion is already closer to metric.

\subsection{Ablations}
\label{sec:exp-ablation}

\textbf{Data mixture.} Table~\ref{tab:data-mixture} varies only the source composition and scores camera-frame hand reconstruction on ARCTIC. Between the two single sources, EgoDex~\cite{hoque2025egodex} recovers the hand more accurately, reaching $48.222$\,mm MPJPE-p, $56.772$\,px EPE-p and $0.097$\,m CT-p, and Ego4D~\cite{grauman2022ego4d} leads only on wrist orientation and smoothness. The mixture is best in five of the six columns and brings MPJPE-p to $26.527$\,mm and CT-p to $0.069$\,m. Since zero-shot transfer is the target, we train on the mixture, whose viewpoints, motions and intrinsics span what the benchmarks present rather than the capture conditions of any one source.

\input{tables/tab_data_mixture}

\textbf{Initialization.} Table~\ref{tab:init} replaces the LingBot-Map~\cite{chen2026lingbotmap} backbone with random initialization and leaves everything else fixed. On camera-frame hands the pretrained backbone is ahead in every column, most visibly on EPE-p, which falls from $82.356$ to $55.038$\,px, and on CT-p, which more than halves. The camera trajectory moves the same way, with relative translation error dropping from $6.130$ to $3.390$\,mm and relative rotation error from $0.846^{\circ}$ to $0.256^{\circ}$. Geometric pretraining therefore supplies the correspondence and scene structure that the hand and camera heads both build on.

\subsection{Inference Throughput}
\label{sec:exp-efficiency}

We evaluate inference efficiency at $512\times384$ and $30$\,fps under identical conditions. \method{} encodes each video window in a single bidirectional spatiotemporal pass. On a single GPU, its per-frame latency is $72.4$\,ms, compared with $105.1$\,ms for HaWoR~\cite{zhang2025hawor} and $1260.0$\,ms for VITRA~\cite{li2025vitra}.

Since the windows are mutually independent, extended egocentric recordings admit parallel processing across GPUs without inter-window coordination, which reduces the mean per-frame latency to $22.7$\,ms. \pipeline{}, although it parallelizes an originally serial cascade, retains a fixed inter-stage dependency whose slowest stage bounds its attainable throughput, and requires $83.4$\,ms per frame. \method{} therefore achieves a $3.67\times$ end-to-end speedup over \pipeline{}, and $12.5\times$ over our reproduction of VITRA, which requires $283.3$\,ms.

\input{tables/tab_efficiency}

%% file: tables/tab_camera_traj.tex
\begin{table}[t]
  \centering
      \caption{World-frame camera trajectory results on the HOT3D~\cite{banerjee2025hot3d} and ARCTIC~\cite{fan2023arctic} validation sets. All sequences are evaluated over their \emph{full} trajectories using $SE(3)$-only alignment without scale fitting, while the trajectory arc-length ratio measures scale consistency. \method{} w/o stage~2 never sees metric ground truth.}
  \label{tab:camera-traj}
  \scriptsize
  \setlength{\tabcolsep}{6.5pt}
  \renewcommand{\arraystretch}{1.1}
  \begin{tabular}{@{}l@{\hspace{4pt}}l rr rr c@{}}
    \toprule
    & Method & \multicolumn{2}{c}{RPE-T $\downarrow$ (mm)}
             & \multicolumn{2}{c}{RPE-R $\downarrow$ (deg)}
             & \multicolumn{1}{c}{Arc len.} \\
    \cmidrule(lr){3-4}\cmidrule(lr){5-6}\cmidrule(l){7-7}
    & & \multicolumn{1}{c}{mean} & \multicolumn{1}{c}{med.}
      & \multicolumn{1}{c}{mean} & \multicolumn{1}{c}{med.}
      & \multicolumn{1}{c}{ratio $\rightarrow 1$} \\
    \midrule
    \multirow{6}{*}[-2pt]{\rotatebox[origin=c]{90}{HOT3D}}
      & DROID-SLAM~\cite{teed2021droidslam} & 5.362 & 3.524 & 0.227 & 0.146 & 0.778 \\
      & InfiniteVGGT~\cite{yuan2026infinitevggt} & 13.524 & 9.673 & 1.492 & 0.392 & 0.556 \\
      & LingBot-Map~\cite{chen2026lingbotmap} & 7.566 & 6.185 & 0.684 & 0.253 & 0.712 \\
      & MegaSaM~\cite{li2025megasam} & \textbf{3.187} & \textbf{2.134} & \textbf{0.082} & \textbf{0.063} & 0.716 \\
    \cmidrule(l){2-7}
      & \method{} w/o stage~2 & 8.752 & 8.371 & 0.234 & 0.229 & 0.466 \\
      & \method{} & 4.694 & 4.783 & 0.284 & 0.259 & \textbf{1.094} \\
    \midrule
    \multirow{6}{*}[-2pt]{\rotatebox[origin=c]{90}{ARCTIC}}
      & DROID-SLAM~\cite{teed2021droidslam} & 33.845 & 14.256 & 1.006 & 0.423 & \textbf{0.964} \\
      & InfiniteVGGT~\cite{yuan2026infinitevggt} & 16.215 & 12.637 & 1.265 & 0.655 & 0.284 \\
      & LingBot-Map~\cite{chen2026lingbotmap} & 9.171 & 8.472 & 0.980 & 0.717 & 0.591 \\
      & MegaSaM~\cite{li2025megasam} & 8.738 & 5.581 & 0.779 & 0.725 & 1.956 \\
    \cmidrule(l){2-7}
      & \method{} w/o stage~2 & 3.536 & 3.624 & \textbf{0.251} & \textbf{0.243} & 0.755 \\
      & \method{} & \textbf{3.392} & \textbf{3.372} & 0.256 & 0.251 & 1.412 \\
    \bottomrule
  \end{tabular}
\end{table}

%% file: tables/tab_init.tex
\begin{table*}[t]
  \centering
  \caption{\textbf{Backbone initialization} on ARCTIC~\cite{fan2023arctic}, for camera-frame hands and for the camera trajectory. Units follow Table~\ref{tab:camera-hand}; RPE-T is in mm, RPE-R in degrees and the arc-length ratio is best at $1$. Best per column in bold.}
  \label{tab:init}
  \scriptsize
  \setlength{\tabcolsep}{3.8pt}
  \renewcommand{\arraystretch}{0.95}
  \begin{tabular*}{\textwidth}{@{\extracolsep{\fill}}lccccccccccc@{}}
    \toprule
    & \multicolumn{6}{c}{Camera-frame hands} & \multicolumn{5}{c}{Camera trajectory} \\
    \cmidrule(lr){2-7}\cmidrule(lr){8-12}
    Init. & MPJPE-p $\downarrow$ & PA-MPJPE-p $\downarrow$ & EPE-p $\downarrow$ & GO-p $\downarrow$ & CT-p $\downarrow$ & Jitter $\downarrow$ & RPE-T mean $\downarrow$ & RPE-T med. $\downarrow$ & RPE-R mean $\downarrow$ & RPE-R med. $\downarrow$ & Arc len. \\
    \midrule
    Random & 48.045 & 26.869 & 82.356 & 24.562 & 0.198 & 18.758 & 6.130 & 6.280 & 0.846 & 0.850 & \textbf{1.201} \\
    LingBot-Map~\cite{chen2026lingbotmap} & \textbf{26.527} & \textbf{15.961} & \textbf{55.038} & \textbf{19.332} & \textbf{0.069} & \textbf{12.959} & \textbf{3.390} & \textbf{3.370} & \textbf{0.256} & \textbf{0.251} & 1.412 \\
    \bottomrule
  \end{tabular*}
\end{table*}

%% file: tables/tab_data_mixture.tex
\begin{table}[t]
  \centering
  \caption{\textbf{Data mixture} for camera-frame hand reconstruction on ARCTIC~\cite{fan2023arctic}; only the source composition changes. EPIC-KITCHENS~\cite{damen2018epickitchens} and the single-rig in-house corpus appear only within the mixture.}
  \label{tab:data-mixture}
  \scriptsize
  \setlength{\tabcolsep}{3.5pt}
  \renewcommand{\arraystretch}{0.95}
  \resizebox{\columnwidth}{!}{%
  \begin{tabular}{lcccccc}
    \toprule
    Sources & MPJPE-p $\downarrow$ & PA-MPJPE-p $\downarrow$ & EPE-p $\downarrow$ & GO-p $\downarrow$ & CT-p $\downarrow$ & Jitter $\downarrow$ \\
    \midrule
    Ego4D only & 49.594 & 26.230 & 64.449 & 23.134 & 0.142 & \textbf{9.642} \\
    EgoDex only & 48.222 & 25.675 & 56.772 & 24.102 & 0.097 & 15.783 \\
    \midrule
    Full mixture & \textbf{26.527} & \textbf{15.961} & \textbf{55.038} & \textbf{19.332} & \textbf{0.069} & 12.959 \\
    \bottomrule
  \end{tabular}}
\end{table}

%% file: tables/tab_efficiency.tex
\begin{table}[t]
  \centering
  \caption{Inference efficiency comparison under identical conditions. VITRA~\cite{li2025vitra} is reproduced for reference.}
  \label{tab:efficiency}
  \scriptsize
  \setlength{\tabcolsep}{4pt}
  \renewcommand{\arraystretch}{0.95}
  \begin{tabular*}{0.9\columnwidth}{@{\extracolsep{\fill}}lccc@{}}
    \toprule
    Method & Time (ms/fr.) $\downarrow$ & fps $\uparrow$ & Speedup $\uparrow$ \\
    \midrule
    \multicolumn{4}{@{}l}{\emph{One GPU}} \\
    VITRA~\cite{li2025vitra} & 1260.0 & 0.8 & -- \\
    HaWoR~\cite{zhang2025hawor} & 105.1 & 9.5 & 12.0$\times$ \\
    \method{} & \textbf{72.4} & \textbf{13.8} & \textbf{17.4}$\times$ \\
    \midrule
    \multicolumn{4}{@{}l}{\emph{Four GPUs}} \\
    VITRA~\cite{li2025vitra} & 283.3 & 3.5 & -- \\
    \pipeline{} & 83.4 & 12.0 & 3.4$\times$ \\
    \method{} & \textbf{22.7} & \textbf{44.1} & \textbf{12.5}$\times$ \\
    \bottomrule
  \end{tabular*}
\end{table}

%% file: sections/7_conclusion.tex
\section{Conclusion}
\label{sec:conclusion}

We presented \method{}, a unified feed-forward model that recovers world-space camera and bimanual hand motion from egocentric video. No hand label in its training data is human-annotated. \method{} is trained on $1{,}021$ hours of motion supervision that \pipeline{} extracts automatically from publicly available egocentric video, and a short second stage uses device-captured camera trajectories from a small in-house corpus to correct absolute scale. The resulting model transfers to unseen domains and is competitive zero-shot on both camera-frame hand reconstruction and camera trajectory estimation. These results indicate that automatically reconstructed motion supervision from public video is sufficient to learn generalizable egocentric motion reconstruction.

\paragraph{Limitations and future work}
\method{} remains affected by the scale drift inherited from monocular reconstruction in the supervision pipeline, and the $32$-frame training window leaves long-horizon drift open. Future work will explore more accurate and diverse metric supervision to reduce scale errors and improve long-range trajectory reconstruction.